\documentclass[letterpaper]{article} 
\usepackage{aaai23}  
\usepackage{times}  
\usepackage{helvet}  
\usepackage{courier}  
\usepackage[hyphens]{url}  
\usepackage{graphicx} 
\usepackage{natbib}  
\usepackage{caption} 
\usepackage{amsmath, amssymb}
\usepackage{subcaption}
\usepackage{dsfont}

\usepackage{algorithm}
\usepackage{algorithmic}

\usepackage{newfloat}
\usepackage{listings}
\DeclareCaptionStyle{ruled}{labelfont=normalfont,labelsep=colon,strut=off} 
\floatstyle{ruled}
\newfloat{listing}{tb}{lst}{}
\floatname{listing}{Listing}
\usepackage{fancyvrb}
\usepackage{fvextra}

\newcommand{\Rho}{\mathrm{P}}
\title{Learning of Generalizable and Interpretable Knowledge in Grid-Based Reinforcement Learning Environments}
\author {
    Manuel Eberhardinger\textsuperscript{\rm 1,2},
    Johannes Maucher\textsuperscript{\rm 1},
    Setareh Maghsudi\textsuperscript{\rm 2}
}
\affiliations {
    \textsuperscript{\rm 1}Hochschule der Medien Stuttgart, Germany\\
    \textsuperscript{\rm 2}University of Tübingen, Germany\\
    eberhardinger@hdm-stuttgart.de, maucher@hdm-stuttgart.de, setareh.maghsudi@uni-tuebingen.de
}

\begin{document}

\maketitle
\begin{abstract}
Understanding the interactions of agents trained with deep reinforcement learning is crucial for deploying agents in games or the real world. In the former, unreasonable actions confuse players. In the latter, that effect is even more significant, as unexpected behavior cause accidents with potentially grave and long-lasting consequences for the involved individuals.
In this work, we propose using program synthesis to imitate reinforcement learning policies after seeing a trajectory of the action sequence. Programs have the advantage that they are inherently interpretable and verifiable for correctness. We adapt the state-of-the-art program synthesis system DreamCoder for learning concepts in grid-based environments, specifically, a navigation task and two miniature versions of Atari games, Space Invaders and Asterix. By inspecting the generated libraries, we can make inferences about the concepts the black-box agent has learned and better understand the agent's behavior. We achieve the same by visualizing the agent's decision-making process for the imitated sequences. We evaluate our approach with different types of program synthesizers based on a search-only method, a neural-guided search, and a language model fine-tuned on code. 
\end{abstract}
\section{Introduction}
Humans can easily explain other agents' behavior, living or artificial, using a single demonstration. However, generating explanations post-hoc in (deep) reinforcement learning (RL) after observing an agent's interactions in an environment remains underexplored. Moreover, it is challenging to produce an informative explanation to understand the agent's reasoning for selecting a specific action for a particular state.

Nevertheless, understanding the behavior of artificial agents is crucial for deploying agents in the real world or games. In games, one wants to ensure the agent behaves similarly and avoid confusing the players with unreasonable actions. In real-world scenarios, this is even more important because, for example, in self-driving cars unpredictable actions can cause accidents and lead to serious harm for those involved. Therefore, RL is not yet applicable in real-world scenarios since the behavior of agents trained with RL is not always predictable and, thus, cannot be verified for all edge cases \cite{dulac-arnold_challenges_2021}.   

In this work, we propose using program synthesis to imitate RL policies after seeing a trajectory of the action sequence. 
Program synthesis is the task of finding a program for a given specification, such as a natural language description or input-output examples \cite{gulwani_program_2017}. 
By distilling neural network policies into programmatic policies, we are able to verify the program for correctness and use traditional formal verification tools \cite{lathouwers_modelling_2022} to analyze the behavior and edge cases of synthesized programs. 
Another benefit of distilling policies into programs is that software developers can adapt the policy to their own needs, which makes it easy to further improve the programmatic policies or adapt them to other scenarios \cite{trivedi_learning_2022}. 

Ideally, we desire to extract programs that can explain decisions and solve the environment; Nevertheless, in this work, we start by dividing complete trajectories into sub-trajectories to be able to find programs at all. Therefore, we intend to lay the foundation for a complete policy extraction algorithm. 

To accomplish this, we adopt a state-of-the-art program synthesis system DreamCoder \cite{ellis_dreamcoder_2021} for learning concepts in grid-based RL environments and demonstrate that it can extract a library of functions. We collect trajectories of the policy, i.e., state-action pairs, from an oracle trained with RL and use the collected data for DreamCoder to imitate these state-action sequences with programs. We use these programs to extract core concepts from the environment, represented as functions. To enable the system to learn a library, we introduce a domain-agnostic curriculum based on the length of the state-action sequences to imitate. By inspecting the generated library, we can make inferences about the concepts the agent has learned and better understand the agent's behavior. We achieve the same by visualizing the agent's decision-making process for the imitated sequences. We evaluate our approach with three different program synthesizers: a search-only approach, a neural-guided search, and a language model fine-tuned on code. 

Our main contributions are as follows:
\begin{itemize}
    \item Introducing a framework for learning reusable and interpretable knowledge that can reason about agent behavior in grid-based reinforcement learning environments
    \item An evaluation of the method on a navigation task through a maze and on two simplified Atari game environments, Asterix and Space Invaders
    \item A comparison of different program synthesis algorithms, including enumerative search, neural-guided enumerative search, and a fine-tuned language model with and without library learning
    \item An analysis of extracted functions of the generated libraries
    \item We open-source the code to enable further research\footnote{\url{https://github.com/ManuelEberhardinger/ec-rl}}. 
\end{itemize}
\section{Related Work}
\paragraph{Program Synthesis and Library Learning}
Program Synthesis has a long history in the artificial intelligence research community \cite{waldinger_prow_1969, manna_knowledge_1975}. In recent years, many researchers have combined deep learning with program synthesis to make program search more feasible by reducing or guiding the search space \cite{balog_deepcoder_2022, nye_learning_2019, ritchie_neurally-guided_2016, chaudhuri_neurosymbolic_2021}. In contrast to the heuristic-based search algorithms, one can also use language models to synthesize programs from text prompts \cite{li_competition-level_2022, odena_program_2021, wang_codet5_2021, poesia_synchromesh_2022, scholak_picard_2021}. Another promising method is learning a library of functions from previously solved problems. These functions are then reusable in an updated domain-specific language to solve more challenging problems \cite{hewitt_learning_2020, ellis_dreamcoder_2021, ellis_learning_2018, cao_babble_2023, bowers_top-down_2023}.
\paragraph{Explainable Reinforcement Learning}
There exists a variety of methods in the explainable reinforcement learning (XRL) domain. In a recent comprehensive survey \cite{qing_survey_2022}, the authors divide XRL into four explainable categories: model, reward, state and task.
Programmatic policies, where a policy is represented by a program, are part of the model-based explanations \cite{verma_imitation-projected_2019, verma_programmatically_2018, anderson_neurosymbolic_2020, trivedi_learning_2022, qiu_programmatic_2022}. 
Other works in this category synthesize finite state machines to represent policies \cite{inala_synthesizing_2020} or use models based on decision trees \cite{silver_few-shot_2020, bastani_verifiable_2018}. Our method belongs to the same category since we explain sub-trajectories of policies, and our main goal in the future is to extract a program that can represent the full policy.
\section{Background}
In this section we give a brief introduction of the different research topics and concepts, that we combine to learn structured and reusable knowledge in grid-based reinforcement learning environments.
\paragraph{Program and Domain-specific Language} 
This work considers programs defined in a typed domain-specific language (DSL) which is based on the Lisp programming language \cite{mccarthy_recursive_1960}. The primitives, i.e. provided functions and constants, of the DSL are control flows, the actions the agent can use, and also modules to perceive the agent's environment. Since we work with grid environments, the agent's perception consists of modules to determine certain positions on the grid and compare them with the available objects in the environment such as walls, empty cells or game-specific objects. The control flows include if-else statements and Boolean operators to formulate more complex conditions. The DSL is a probabilistic grammar with a uniform distribution over the primitives, i.e., each primitive is assigned the same probability of being used.  Listing \ref{lst:listing} shows an example program that gets an object on the map x and compares it to a wall object. If there is a wall at the specified position, the left action is chosen, otherwise the forward action.
The full DSL is included in Appendix \ref{appendix:dsl}. 
\begin{listing}[t]%
\caption{An example program generated from the DSL.}%
\label{lst:listing}%
\begin{lstlisting}[language=Java, escapeinside={(*}{*)}]
(*$\lambda$*)(x) ( 
  (if (eq-obj? wall-obj (get x 1 0))
        left-action forward-action)
)
\end{lstlisting}
\end{listing}

\paragraph{Program Synthesis}
One of the core component in this work is the program synthesizer. 
Our work is based on DreamCoder, a state-of-the-art program synthesis system that combines program synthesis with library learning. 
DreamCoder provides an implementation of a search algorithm that enumerates programs in decreasing order of their probability of being generated based on a given probabilistic grammar. 
The programs are checked if they fit a given specification until the top $k$ most likely solutions are found or a timeout is reached \cite{ellis_dreamcoder_2021}. 

To improve search time, a neural network is trained to predict a distribution over the programs defined by the DSL, thus, adapting the uniform distribution to the one that fits the training data of the neural network. This means that the network predicts the probability of the primitives in the DSL, which results in programs being found faster because they are checked earlier by the enumerative search algorithm \cite{ellis_dreamcoder_2021}. 
\begin{figure*}[t]
\centering
\includegraphics[width=0.65\textwidth]{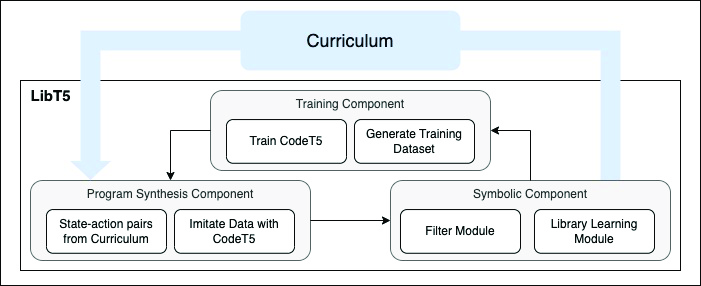}
\caption{The architecture for creating explanations for a given reinforcement learning environment. The framework can be decomposed into three components that are executed iteratively and are guided by a curriculum. First, we train the CodeT5 model, then we synthesize programs for the provided data from the curriculum. Finally, the synthesized
programs are evaluated by the symbolic component for correctness and analyzed to extract a library of functions. We describe the method in detail in Section \ref{method}.}
\label{fig:overview}
\end{figure*}

In order to perform a study between different types of program synthesizers, we replace DreamCoder's program synthesis component with a neural program synthesizer. For this, we use CodeT5 \cite{wang_codet5_2021}, a finetuned T5 model \cite{raffel_exploring_2020} on multiple programming languages and code related tasks. \citet{raffel_exploring_2020} introduced the T5 model with an encoder-decoder architecture and unified different natural language processing (NLP) tasks into a single one by converting them into a text-to-text format. That allows the authors to treat every problem in one way, i.e., using text as input and producing text as output. In this work, CodeT5 is further trained on Lisp programs to synthesize programs in the provided DSL by converting the agent's observation into a text prompt and synthesizing programs in text format as output. 
\paragraph{Library Learning}
The goal of learning libraries is to build a library of specialized concepts in a particular domain that allow programs to be expressed in a concise way. This is similar to software engineers using open source libraries to improve their programming efficiency, since someone else already implemented the needed concepts in a given domain.
We use the DreamCoder library learning module to extract functions from solved tasks by analyzing synthesized programs. These programs are refactored to minimize their description length while growing the library. Instead of only extracting syntactic structures from programs, DreamCoder refactors programs to find recurring semantic patterns in the programs \cite{ellis_dreamcoder_2021}. In this work, we use library learning to build a library of functions for grid-based RL environments that allow us to make inferences about the knowledge acquired by the black box agent during training.

\paragraph{Imitation Learning}
To simplify the problem, we limit ourselves to imitation learning, instead of directly finding programs from rewards \cite{hussein_imitation_2017}. Although directly finding programs from rewards is an interesting challenge for future research, our main objective is to make the agent's behavior interpretable. We define the problem we want to solve as an imitation of sub-trajectories of state-action pairs collected from a previously trained agent. 

\section{Method}
\label{method}
Figure \ref{fig:overview} shows a high-level overview how we adapted DreamCoder with a neural program synthesizer based on the CodeT5 model \cite{wang_codet5_2021}, therefore we call this approach LibT5. The framework consists of three components and a curriculum. In addition, we need an oracle for collecting the data to be imitated. The collected data is provided to the curriculum in the beginning of the training process. For evaluating DreamCoder on the problem of imitating sub-trajectories of state-action sequences, we exchange the LibT5 component with DreamCoder and integrate it into the curriculum. The main differences of LibT5 compared to DreamCoder are the program synthesizer and that DreamCoder first performs an enumerative search before training the neural network and then includes programs found from solved tasks in the training data set. LibT5 is trained only on random programs. 

As the oracle is dependent on the domain and not part of the method, we describe the used oracles in Section \ref{section:domains}.
\subsection{LibT5}
LibT5 consists of three components that are executed iteratively. First, we train the CodeT5 model, then we synthesize programs for the provided test data. Finally, the synthesized programs are evaluated by the symbolic component for correctness and analyzed to extract a library of functions.

\paragraph{Training Component} The first part is the training process of the CodeT5 model \cite{wang_codet5_2021} with randomly generated programs from the current DSL. The randomly generated programs are executed in a given environment for $t$ steps to collect input-output examples, i.e., sequences of state-action pairs to be imitated, as a training dataset. $t$ is chosen randomly for each program, so we do not overfit on a specific sequence length. 

In our setup, we generate 50000 random programs. They are then executed in a randomly selected environment of the provided ones for each domain to collect data to imitate (see Figure \ref{fig:envs} for example environments). We execute each program with a random sequence length $t$ between $t_{min}$ and $t_{max}$. The programs do not have a specific target or reward since they are sampled from the DSL. Our goal in creating a training data set is to exhibit the behavior of programs in a specific RL domain, i.e., how the agent is controlled by given programs in a domain. Random program generation is limited to a maximum depth $d_{max}$ of the abstract syntax tree. We train the model for five epochs in each iteration. 

CodeT5 is used without any modifications, as we generate text prompts from the state-action pairs which the model maps to the random programs. The agent's observation, a 2D array of integers, is converted into a string representation, where each integer represents an object in the environment, such as a wall or an enemy. Then the action is appended after the observation. We explain the text prompt generation in Appendix \ref{appendix:text-prompts}.

\paragraph{Program Synthesis Component} The second component converts the data provided from the curriculum for the current sequence length into the text prompt for the model, and then the model synthesizes $\Rho$ programs for the state-action sequences to be imitated. These programs are then passed to the symbolic part of the framework. 
\paragraph{Symbolic Component} In this component, the filter module evaluates $\Rho$ programs for syntactic and functional correctness in the provided DSL that imitates the state-action sequence.

The library learning module uses the correct programs to generate a library by extracting functions from the found programs and adding them to the current DSL. It extracts functions only if a part in a program occurs multiple times in other synthesized programs on the oracle data. That way, the extracted functions are beneficial for the DSL since they have been synthesized several times for different state-action sequences. 
%
\subsection{Curriculum}
The curriculum is based on the action sequence length and, therefore domain-agnostic. 
We start with an initial sequence length of three, and at each iteration, after the symbolic component has completed the library learning phase, we check whether the sequence length should be increased.
We always sample new random programs from the DSL and run them in the environment as the library is updated each iteration to represent more diverse programs. 
We increment the action sequence length if at least 10\% of the oracle's data is imitated and stop the training process if the action sequence length has not been incremented twice in a row.

This curriculum strategy is based on the assumption that longer sequence lengths are more complex than shorter ones. Programs that need to imitate three actions do not need to represent as much information as programs that imitate five actions; Thus, the program length is shorter. Shorter programs are easier to synthesize compared to long ones because of the smaller search space. \citet{ellis_dreamcoder_2021} showed that building up a library of complex functions, enables DreamCoder to synthesize programs for more difficult tasks. 

Table \ref{table:params-domains} shows the different parameters for the evaluated domains, as they depend on the complexity and observation space of the environment.
\begin{table}
    \centering\resizebox{\linewidth}{!}{%
    \begin{tabular}{ | c || c | c | }
        \hline
         Parameters & Navigation Task (5x5) & MinAtar Games (10x10) \\ \hline
         $t_{min}$ & 5 & 3 \\ 
         $t_{max}$  & 60 & 20 \\
         $d_{max}$  & 6 & 20 \\
         $\Rho$ & 100 & 500 \\
         \hline
    \end{tabular}}
    \caption{The parameters for the different domains. MinAtar games have a grid size of 10x10, while the navigation task has a partial observation of size 5x5.}
    \label{table:params-domains}
\end{table}
\section{Experiments}
In this section we evaluate the different program synthesis systems on the problem of imitating sub-trajectories on two different domains. 
We first introduce both domains. 
Then, we evaluate the conducted experiments, followed by an introduction of the method for generating explanations from programs. Finally, we perform a thorough analysis of the extracted functions in the library.
\subsection{Domains}
\label{section:domains}
\begin{figure}[t]
\begin{subfigure}[b]{1\linewidth}
  \centering
  \includegraphics[width=\linewidth]{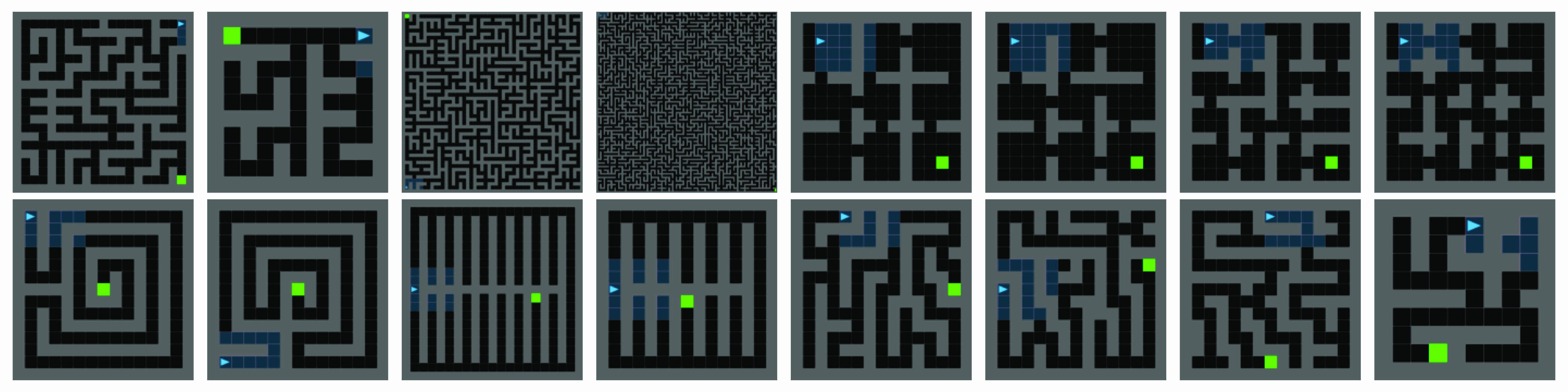}
  \caption{Gridworld: The different environments used for generating the training data. The evaluation is always performed on the medium sized perfect maze environment which is placed on the top left.}
    \label{fig:minigrid}
\end{subfigure} 
\begin{subfigure}[b]{\linewidth}
\centering
\includegraphics[width=.25\textwidth]{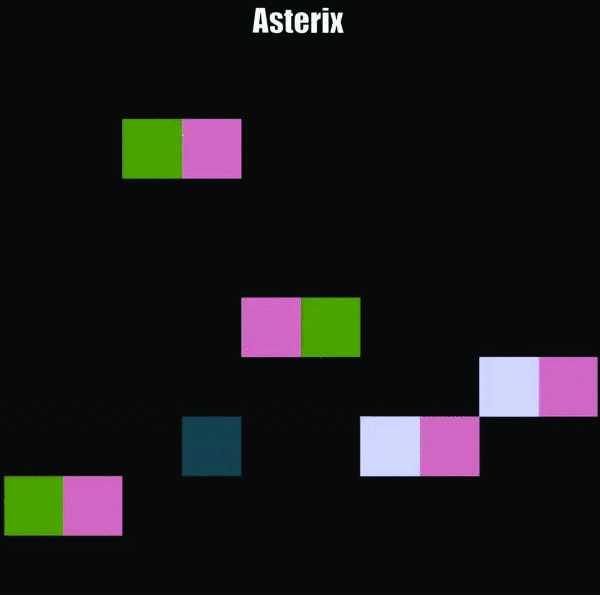}
\includegraphics[width=.25\textwidth]{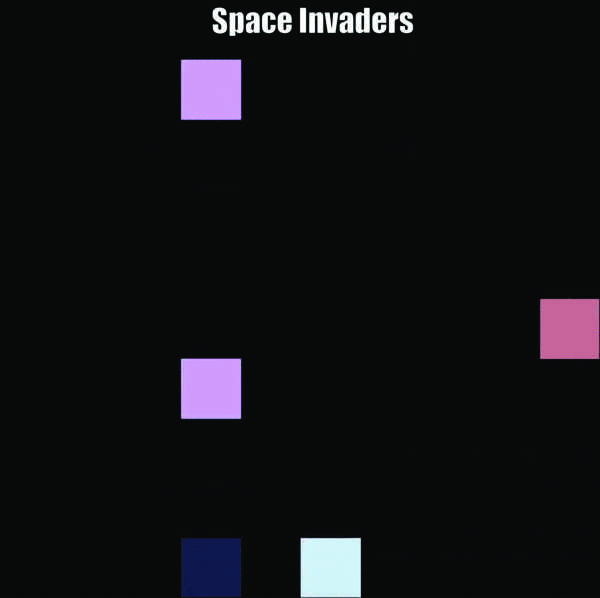}
\caption{MinAtar: The two different miniaturized versions of the Atari 2600 games, Asterix and Space Invaders. In all environments the agent is displayed in dark blue. The other colors are game-specific objects, such as gold coins, enemies and bullets in Asterix or Space Invaders (from \cite{young_minatar_2019}).}
\label{fig:minatar}
\end{subfigure} 
\caption{The environments for the Gridworld and the MinAtar domain used for the training data generation. }
\label{fig:envs}
\end{figure}

\paragraph{Gridworld} The first domain we evaluate on is a navigation task from the grid-world domain \cite{hevalier-boisvert_maxime_minimalistic_2018}.
We trained the agent with the default hyperparameters from \citet{parker-holder_evolving_2022} and then collected state-action pairs from the medium-sized perfect grid environment displayed on the top left in Figure \ref{fig:envs}. 

\paragraph{MinAtar}
\citet{young_minatar_2019} introduced a miniature version for five games of the Arcade Learning Environment \cite{bellemare_arcade_2013}. The games are simplified to enable more efficient experimentation. MinAtar converted the game into a symbolic 10x10 grid representation with the same game dynamics. Each environment provides a one-hot encoded 10x10x$n$ observation, where $n$ channels correspond to specific game objects such as cannon, enemy or bullet objects in Space Invaders. For our experiments we converted the state into a single 10x10 grid. We evaluate our method on Asterix and Space Invaders. We trained both agents with the default parameters provided from \citet{young_minatar_2019}.

To generate diverse training data that starts not always from similar game states, we let the oracle play the episode for a random time before executing the program. This ensures that the training data set is capturing different aspects of the policy. 

\subsection{Evaluation}
We evaluate the problem of imitating sub-trajectories for RL environments with four different program synthesizers:

\begin{itemize}
    \item Search: Program synthesis with an enumerative search algorithm. We use the implementation from \citet{ellis_dreamcoder_2021} and the same DSL as DreamCoder to show the benefits of the neural-guided search.
    \item DreamCoder: A neural-guided search algorithm with a library learning module \cite{ellis_dreamcoder_2021}.
    \item CodeT5: A language model fine-tuned on Lisp programs on our data \cite{wang_codet5_2021}.
    \item LibT5: The CodeT5 model combined with DreamCoder's library learning module.
\end{itemize}
For the final evaluation, we use data collected from the same agent but on different runs to ensure that we do not evaluate and train on the same data. The performance is measured by

$$
Accuracy = \frac{1}{N}\sum_{\tau \in D} f(\Rho, \tau),
$$
$$
f(\Rho, \tau) = 
\begin{cases}
    1,& \text{if } \sum_{\rho \in \Rho} g(\rho,\tau) > 0\\
    0,              & \text{otherwise}
\end{cases}
$$
$$
g(\rho,\tau) = { \underbrace{ \mathds{1} \left\{ \text{ }\operatorname{EXEC}(\rho,s)==a,  \text{         } \forall (s,a) \in \tau \text{     }\right\}}_{\text{is } 0 \text{ after the first } (\rho,s) \text{ where } \operatorname{EXEC}(\rho, s) \text{ } != \text{ }  a}} 
$$
where $N$ is the size of the dataset $D$ to imitate, $\tau$ is a sub-trajectory from $D$ that consists of state-action pairs $(s,a)$, and $\Rho$ is the set of all synthesized programs from a given method. 
$f(\Rho, \tau)$ checks if there exists any program $\rho$ out of all synthesized programs $\Rho$ that is correct.
$g(\rho,\tau)$ evaluates if a given program $\rho$ can imitate the full rollout $\tau$ and returns 1 if this is the case and otherwise 0.
$\operatorname{EXEC}(\rho, s)$ executes the program on a given state $s$  and returns an action $a$.
The identity function $\mathds{1}$ maps Boolean values to 0 and 1.

Since we have formulated the problem as an imitation of sub-trajectories, we cannot use a more appropriate metric for evaluation. In RL, there are often many different trajectories to achieve a goal in an environment, but in our case we need to evaluate our framework using a more rigorous metric until a complete policy can be distilled into a program with our framework.

\paragraph{Fairness of evaluation} 
Considering fundamental differences, a fair comparison of the used algorithms can be challenging. We describe the used hardware resources and the main distinctions of the experimental setup in more detail in Appendix \ref{appendix:setup}.

\begin{figure}
    \begin{subfigure}[b]{\linewidth}
      \centering
      \includegraphics[width=\linewidth]{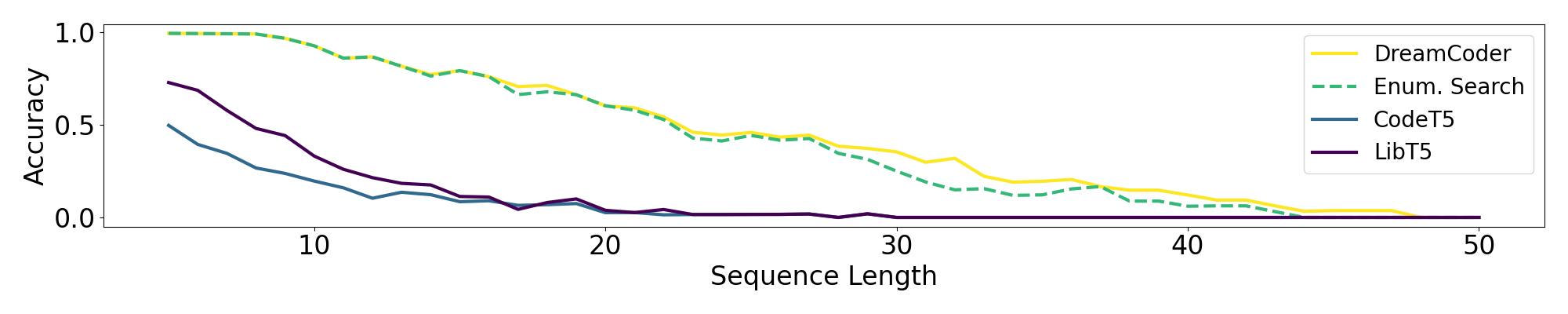}
      \caption{The perfect maze environment.}
      \label{fig:maze-eval}
    \end{subfigure}
    \begin{subfigure}[b]{\linewidth}
      \centering
       \includegraphics[width=\linewidth]{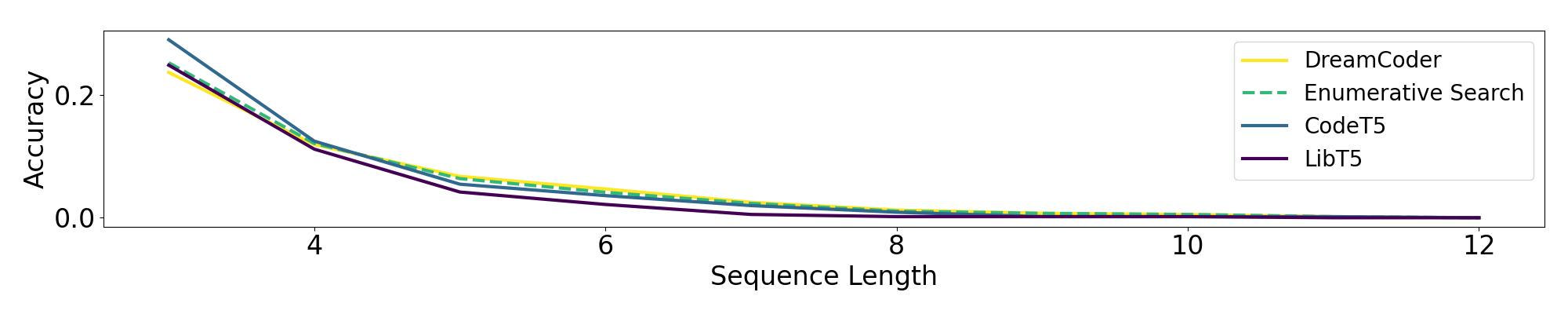}
       \caption{The Asterix environment.}
       \label{fig:asterix-eval} 
    \end{subfigure} 
    \begin{subfigure}[b]{\linewidth}
      \centering
       \includegraphics[width=\linewidth]{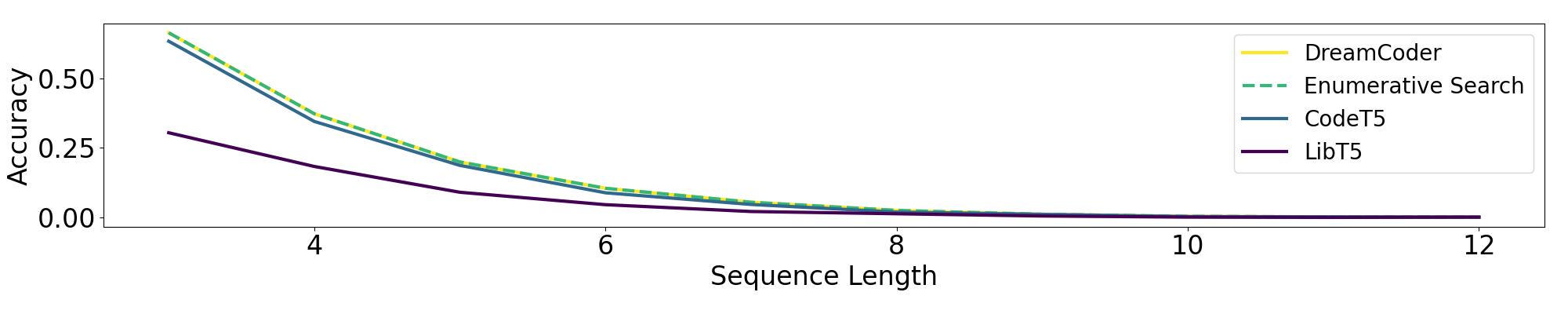}
       \caption{The Space Invaders environment.}
       \label{fig:space-invaders-eval} 
    \end{subfigure} 
    \caption{The evaluation of the different methods on the three environments. The evaluation data was collected on new rollouts of the trained agent. We evaluated the percentage of the correct imitated sub-trajectories for an increasing sequence length until no more programs were found.}
    \label{fig:eval}
\end{figure}

\paragraph{Results} Figure \ref{fig:eval} shows the final evaluation of newly collected data in the same environment used to extract functions from found programs on the solved test tasks. Depending on the domain, different program synthesis methods are well suited. Figure \ref{fig:maze-eval} shows the evaluation for the maze environment with the smallest observation space of 5x5. The search-based methods can solve sub-trajectories almost twice as long as the neural-based models. For Asterix all methods show similar performance (Fig. \ref{fig:asterix-eval}). For Space Invaders, LibT5 performs worse compared to the other methods (Fig. \ref{fig:space-invaders-eval}).

Figure \ref{fig:maze-eval} shows that library learning can be useful for neural program synthesizers, but also detrimental depending on the environment. For both MinAtar environments, LibT5 is not as good as CodeT5. This suggests that the more diverse DSL can lead to the problem of  ``catastrophic forgetting" \cite{kirkpatrick_overcoming_2017}, and previously solved programs become unsolvable. In addition, longer action sequences are no longer solved as well. Our hypothesis is that the library is not beneficial, although more functions have been extracted from LibT5 compared to DreamCoder (see Table \ref{table:num-functions}). Inspecting programs synthesized with LibT5 for Space Invaders shows that they are too complicated compared to programs synthesized with CodeT5. The reason is that LibT5 uses the extracted functions of the library, even if the task is easier to solve with the initial primitives. 

From both observations, we conclude that neural program synthesizers may be useful for larger observation spaces. Catastrophic forgetting could be mitigated by adjusting the probabilities of the functions in the probabilistic grammar according to their usefulness. By lowering the probability of the more complex functions, the grammar will produce simpler programs. In addition, we need to improve the generation of training data by collecting different runs for each program or trying different representations for encoding  the state-action pairs. Therefore, further research is imperative to better integrate the library learning module into the framework with neural program synthesizers.

It is also evident from Figure \ref{fig:eval} that a system without a curriculum cannot imitate complete action sequences, as it can currently imitate up to sequence lengths of 50. In comparison, complete trajectories are up to 1000 steps long depending on the environment.

\subsection{Inspecting the Program Library}
\begin{figure*}
  \centering
  \includegraphics[width=\linewidth]{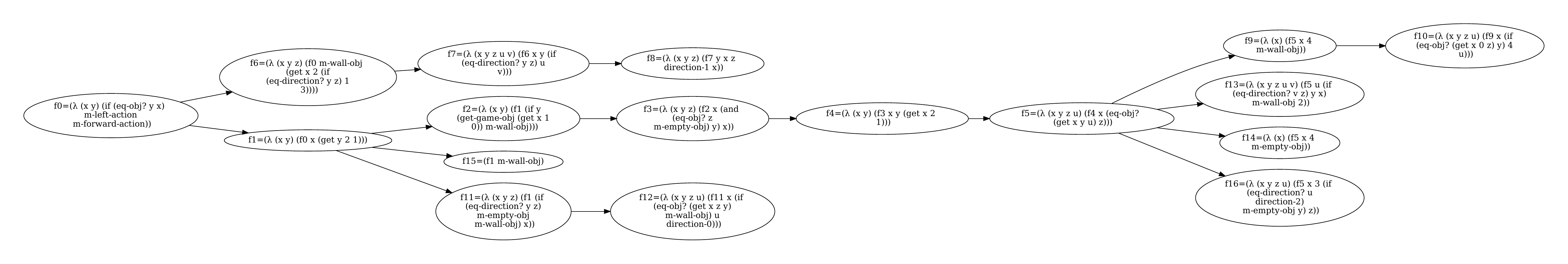}
  \caption{Maze: The extracted functions from programs found by using DreamCoder. The function \texttt{f10} is using seven previous discovered functions (zoom in for better visibility).}
    \label{fig:ec-maze-lib}
\end{figure*}
\begin{table}[b]
\centering
\resizebox{0.6\linewidth}{!}{%
\begin{tabular}{ | c || c | c | }
\hline 
Environment & DreamCoder & LibT5 \\
\hline
Maze & 17 & 27 \\
Asterix & 4 & 8 \\
Space Invaders & 15 & 23 \\
\hline
\end{tabular}}
\caption{Number of extracted functions for different program synthesis methods.}
\label{table:num-functions} 
 \end{table}
In this section, we analyze the libraries extracted from the evaluated methods. Appendix \ref{appendix:libs} includes the full libraries. Table \ref{table:num-functions} shows the number of extracted functions for the different program synthesis methods. LibT5 extracts for all environments the most functions. 

\begin{figure}
  \centering
  \includegraphics[width=\linewidth]{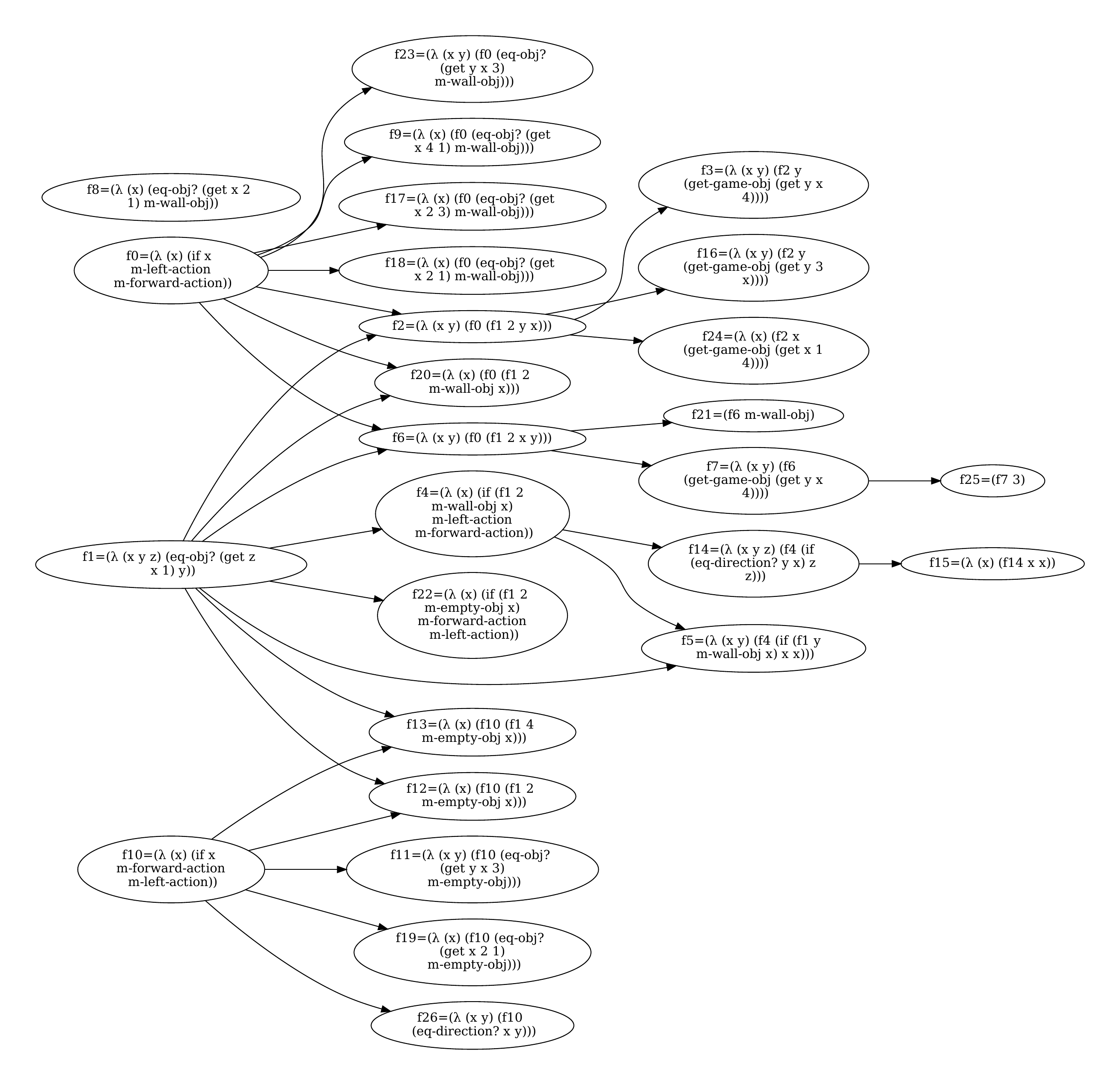}
  \caption{Maze: The extracted functions from programs found by using LibT5 (zoom in for better visibility).}
    \label{fig:t5-maze-lib}
\end{figure}
Figure \ref{fig:ec-maze-lib} shows the discovered library of DreamCoder for the maze environment with a deep hierarchical structure. \verb_f10_ is using seven previous discovered functions. Figure \ref{fig:t5-maze-lib} shows the library of LibT5. In contrast to DreamCoder, more functions were extracted, but more often semantic similar functions are found and the deep hierarchical structure is missing. The use of a language model for program search in combination with library learning raises a new problem similar to one previously addressed in Inductive Logic Programming. \citet{cropper_forgetting_2020} analyzed what the perfect library size is and how to forget unnecessary programs in the library. This is also necessary in our case, as we assume that LibT5 synthesizes many programs that are semantically the same but differ syntactically. Therefore, the library learning module extracts many similar functions and adds them to the library. A similar problem is also observable in the AlphaCode system, which clusters synthesized programs before selecting solutions for submission to programming competitions \cite{li_competition-level_2022}. From this we conclude, that a larger library is not always beneficial for the program synthesizer. 

\subsection{Visualization of the Decision Making Process}
\label{visualization-process}
\begin{figure*}[t]
\centering
    \includegraphics[width=\linewidth]{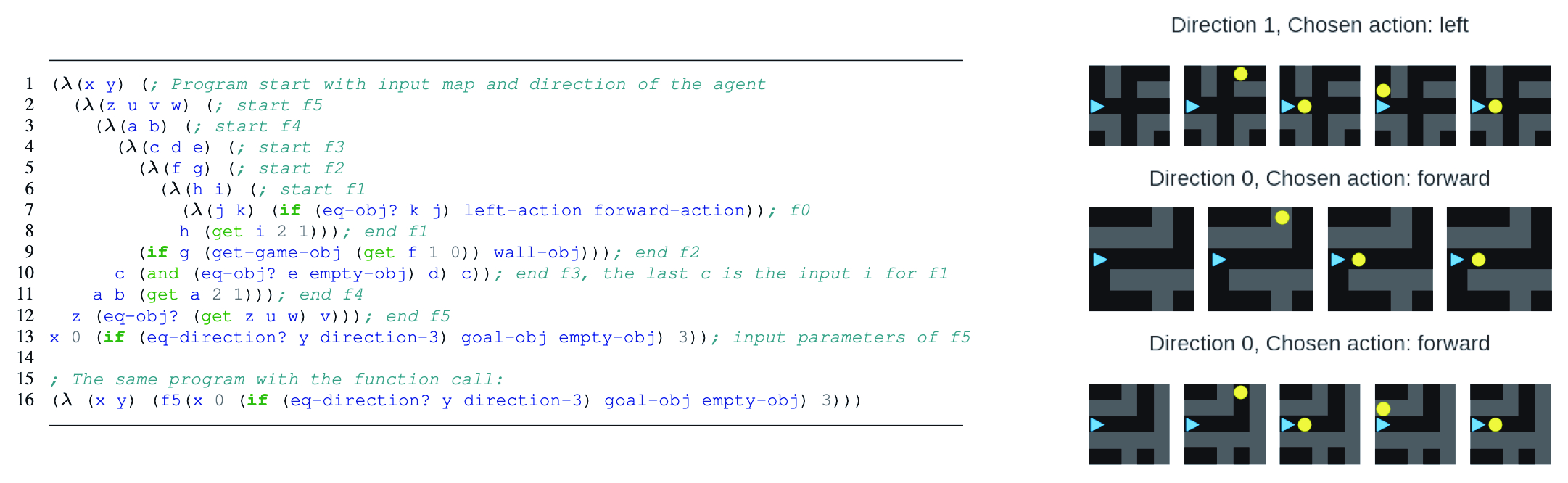}
  \caption{
  Left: The program for a given sub-trajectory synthesized by DreamCoder. Line 1 to 13 show the program with the implementation of the functions. Line 16 shows the same program when calling \texttt{f5} from the library. Right: The decision-making process, when executing the program on the state-action sequence. We show explanations for three of 24 states of a given sub-trajectory. The grid positions that are checked in the program are yellow. The agent's position is marked blue and faces to the right. Grey and black indicate walls and empty cells, respectively. The forward action moves the agent one grid cell to the right. The left action only turns the agent $90^{\circ} $ in the left direction but does not move it. The cell on position (2,1) is checked multiple times, as at first it is checked in \texttt{f4} and then later in \texttt{f1}. We give a detailed explanation of the program in Section \ref{visualization-process}.
  }
    \label{fig:vis-process}
\end{figure*}
Since programs are inherently interpretable, we developed a method to visualize the agent's decision-making process by highlighting those grid positions responsible for choosing a particular action. Since one position is not always sufficient to select the correct action, we create step-by-step explanations of the  ``reasoning process" by traversing the program call graph \cite{ryder_constructing_1979} and logging all function calls and their parameters. 

Figure \ref{fig:vis-process}-left shows a program synthesized by DreamCoder. The data was collected from the maze environment for a sub-trajectory of 24 state-action pairs. From line 1 to 13, we show the full implementation of the program. The first $\lambda$ denotes the start of the program and receives two input parameters, the map and the direction of the agent. All subsequent $\lambda$ represent discovered functions from the library, followed by the input parameters. In line 16, we show the program when we use the function \verb_f5_ from the library. This shows the effectiveness of using library learning in combination with program synthesis. 

Figure \ref{fig:vis-process}-right visualizes three examples of the reasoning process by highlighting the responsible grid cells in yellow. The agent's position is in blue, which is the same in all visualizations because the partial observation of the agent is aligned to the same position by the environment. The direction above the images indicates in which direction the agent is looking on the full map, since the surrounding area is only partially visible. The walls are gray, and the path through the maze is black. The program first checks if the direction is \verb_direction-3_ and returns an \verb_goal-object_ or \verb_empty-object_. Since for all three visualizations the direction is one or zero, the \verb_empty-obj_ is always returned. Then the \verb_empty-obj_ is compared with the cell on position (x=3,y=0). The result is an input parameter for \verb_f4_ and \verb_f3_. The coordinates (3,0) are also input parameters of \verb_f5_. Then \verb_f4_ is called which gets the object on position (1, 2). This object is then compared with an empty object in \verb_f3_ and it is checked whether the other input parameter of \verb_f5_ is also true. Depending on this result, the object at position (0,1) or a \verb_wall-obj_ is returned. The returned object is then compared to the position (1,2) and finally the agent decides whether to use the left or the forward action.

Currently, the explanations of the policy can be incorrect, as we do not have a complete policy extraction algorithm and only imitate sub-trajectories collected from an oracle. Without imitating the complete trajectories, the created explanations can be wrong; As such, the programs found for longer action sequences are more reliable as they explain more of the policy. 

\subsection{Discussion \& Limitations}
In our experiments, we have demonstrated that DreamCoder and LibT5 are able to learn a library of functions for a navigation task and two game environments with a discrete state and action space. By traversing the program call graph of synthesized programs, we created visual explanations for the agent's decision-making process. We concluded our experiments with an analysis of the generated libraries for the given domains and discussed the implications. 

While learning a library of concepts showed promising results for grid-based environments with a small observation space, we need to further improve our framework for medium-sized and large observation spaces. We have shown that it is possible for the MinAtar environments to learn a library and imitate short sequences using the CodeT5 model, but the library was not used effectively and the synthesized programs were too complicated compared to the data to be imitated. Therefore, we need to further investigate how the library learning module can benefit neural program synthesizers without compromising its ability to imitate shorter state-action sequences. If this is possible, this opens up other interesting domains, such as AlphaGo \cite{silver_alphago_2016}, where humans struggle to comprehend the reasoning process of strategies discovered by artificial agents through self-play.

Additionally, for these environments, it is straightforward to define the functional primitives for the agent's perceptions and actions. However, that becomes challenging for continuous- state and action spaces, or when an image represents the state. For images, we could use an object detection model which parses the images before generating text prompts for the program synthesizer, similar to \cite{yi_neural-symbolic_2018}, where an image is parsed into a structural representation that is then used in a program.  For continuous representations, further research is imperative to verify the effectiveness of this method.
\section{Conclusion and Future Work}
In this paper, we adapted the DreamCoder system to learn structured and reusable knowledge in grid-based reinforcement learning environments that allows reasoning about the behavior of black-box agents. We further evaluated the use of a neural network as a program synthesizer and discussed the positive and negative aspects of both methods. The main disadvantage of the proposed framework is its dependence on an oracle for collecting trajectories, whereas it does not depend on much background knowledge except for the initial primitives in the DSL. 

This work opens many possibilities for future work. The main focus is a policy extraction algorithm that can imitate the entire state-action sequences and not only parts of them. Additionally, we want to evaluate our method on continuous or image-based domains to validate that it is domain-agnostic. 

\appendix
\section{Domain-specific Language}
\label{appendix:dsl}
Table \ref{table:dsl} shows the initial domain-specific language, which contains only the primitives necessary to get different cells on the grid, the control flow structures and Boolean operators. Since we use a typed DSL, we show the types for each function or value. If our primitive is a value, only one type appears in the type column. For functions, multiple types are combined with an arrow $\small\rightarrow$. The last type represents the return value of the function. The types before it are the types of the input parameters. The type \verb_func_ represents a function because if-clauses returns a new function to execute depending on the condition since partial programs are also functions in Lisp.

To generate random programs, we can specify the types of program to be generated. In our case, we always want programs of type \verb_map_ $\small\rightarrow$ \verb_direction_ $\small\rightarrow$ \verb_action_ or \verb_map_ $\small\rightarrow$ \verb_action_, so a random program is always defined from two input parameters of type \verb_map_ and \verb_direction_ or one input parameter \verb_map_ and returns a value of type \verb_action_. We restrict the input parameter types of the \verb_eq-obj?_ function to \verb_mapObject_ and \verb_object_ so that sampling programs from the grammar always results in the comparison of at least one object from the map.
\begin{table*}
    \centering
    \resizebox{\linewidth}{!}{
    \begin{tabular}{llll}
         Primitive Function/Values & Description & Type & Note \\ \hline
         left, right, forward, down, fire, no-op & possible actions & action &  \\ 
         0, 1, ...,  9 & integer values & int & maze: 0,..,5\\
         2D array & 2D grid observation of the agent & map \\
         direction-0, direction-1 & represents either north, east, south and west & direction & only for maze \\ 
         direction-2, direction-3 & & &  \\
         depends on the environment & possible objects on the map & object \\
         depends on the environment & a object on a map with (x, y)-coordinates & mapObject \\
         if & standard if-clause & bool $\small\rightarrow$ func $\small\rightarrow$ func $\small\rightarrow$ func\\
         eq-direction? & checks if two directions are equal & direction $\small\rightarrow$ direction $\small\rightarrow$ bool& only for maze \\
         eq-obj? &  checks if two objects are equal & mapObject $\small\rightarrow$ object $\small\rightarrow$ bool \\
         get & get a object on the map for two coordinates & map $\small\rightarrow$ int $\small\rightarrow$ int $\small\rightarrow$ mapObject \\
         get-game-obj & get the object type of a mapObject & mapObject $\small\rightarrow$ object  \\
         not & negates a Boolean value & bool $\small\rightarrow$ bool \\
         and &  conjunction of two Boolean values & bool $\small\rightarrow$ bool $\small\rightarrow$ bool \\
         or &  disjunction of two Boolean values & bool $\small\rightarrow$ bool $\small\rightarrow$ bool \\
         get-x & get the X coordinate of a mapObject & mapObject $\small\rightarrow$ tx  \\
         get-y &  get the Y coordinate of a mapObject  & mapObject $\small\rightarrow$ ty \\
         eq-x? &  checks if two X coordinates are the same & tx $\small\rightarrow$ tx $\small\rightarrow$ bool \\
         eq-y? &  checks if two Y coordinates are the same & ty $\small\rightarrow$ ty $\small\rightarrow$ bool \\
         gt-x? &  checks if the first X is greater than the second & tx $\small\rightarrow$ tx $\small\rightarrow$ bool \\
         gt-y? &  checks if the first Y is greater than the second & ty $\small\rightarrow$ ty $\small\rightarrow$ bool \\
    \end{tabular}}
        \caption{The used domain-specific language at the beginning. The type column shows one type for values and several types separated by an arrow for functions. The type after the last arrow is the return type of the function. The types before it are the types of the input parameters. }
    \label{table:dsl}
\end{table*}
\section{Text Prompts}
\label{appendix:text-prompts}
Text prompts are generated by converting the agent's observation into a string representation and then concatenating the string representation with the action. This is repeated for all state-action pairs until each pair in the sequence is represented as a string. Then all strings are combined into a single text prompt.

\begin{figure}
  \centering
  \includegraphics[width=\linewidth]{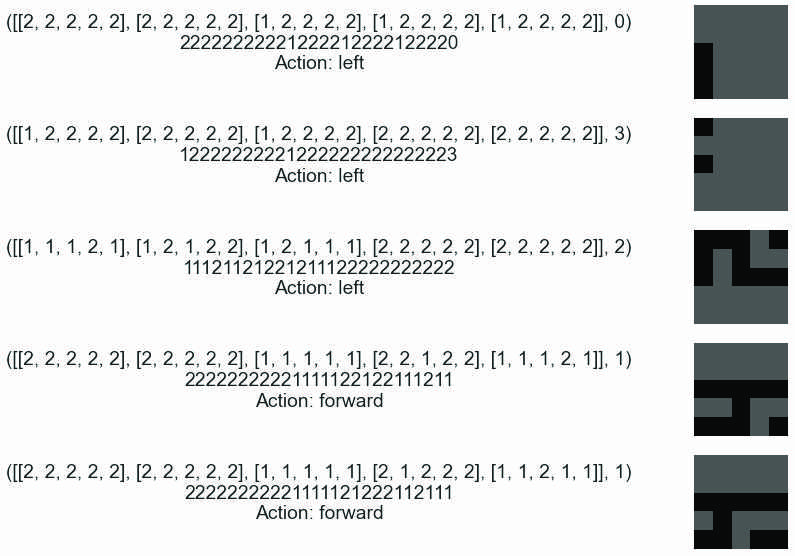}
  \caption{Text prompts are created by converting the 2D array into a string were all values are concatenated without spaces.}
    \label{fig:text-prompts}
\end{figure}

Figure \ref{fig:text-prompts} shows an example of a state-action sequence of length five. On the right of the first line is the 2D array, followed by the string representation and the selected action. On the left is the corresponding maze represented by the 2D array.
The final text prompt is then generated by iteratively concatenating all the string representations and actions for the entire sequence. The final text prompt for the state-action sequence is: 
\begin{Verbatim}[breaklines=true]
22222222221222212222122220 left 12222222221222222222222223 left 
11121121221211122222222222 left 22222222221111122122111211 forward 
22222222221111121222112111 forward
\end{Verbatim}
The $1$ represents empty grid cells and the $2$ represents wall objects on the map. 

\section{Experimental Setup \& Hardware Resources}
\label{appendix:setup}
Table \ref{table:params} shows the hyperparameters for the different methods. We use the same hyperparameters for each iteration. We adopted the hyperparameters from the DreamCoder system to our problem and tried out a few timeouts; since our system is executed iteratively, the runtimes of the experiments are very long, so an extensive hyperparameter search was not possible.
For the neural-guided search, we train an encoder-decoder neural network. The encoder for the state observations of the maze was adapted from \citet{parker-holder_evolving_2022}. For the MinAtar environments we adapted the encoder from \citet{young_minatar_2019}. The decoder, which predicts which functional primitives are used in the program to be synthesized, is generated automatically from the DreamCoder system \cite{ellis_dreamcoder_2021}. We train the neural-guided search model for 5000 update steps. \citet{ellis_dreamcoder_2021} explain that DreamCoder does not require a large data set because it is based on a search method that is already defined by the DSL. The neural network only improves the search. CodeT5, on the other hand, learns everything from data and therefore requires many more programs. The library learning module is restricted to an arity of three, which means that extracted functions can have up to three input parameters.

\begin{table}
    \centering\resizebox{\linewidth}{!}{%
    \begin{tabular}{lllll}
         Hyperparameters & Enum. Search & DreamCoder & CodeT5 & LibT5 \\ \hline
         Search Timeout & 720 seconds & 720 seconds &  - & - \\ 
         Training time & - & 5000 steps & 5 epochs & 5 epochs \\
         Training Programs & - & 5000 & 50000 & 50000 \\
         Arity & 3 & 3 & - & 3\\
    \end{tabular}}
    \caption{The hyperparameters for the different methods.}
    \label{table:params}
\end{table}

\section{Extracted Libraries}
\begin{figure}
  \centering
  \includegraphics[width=\linewidth]{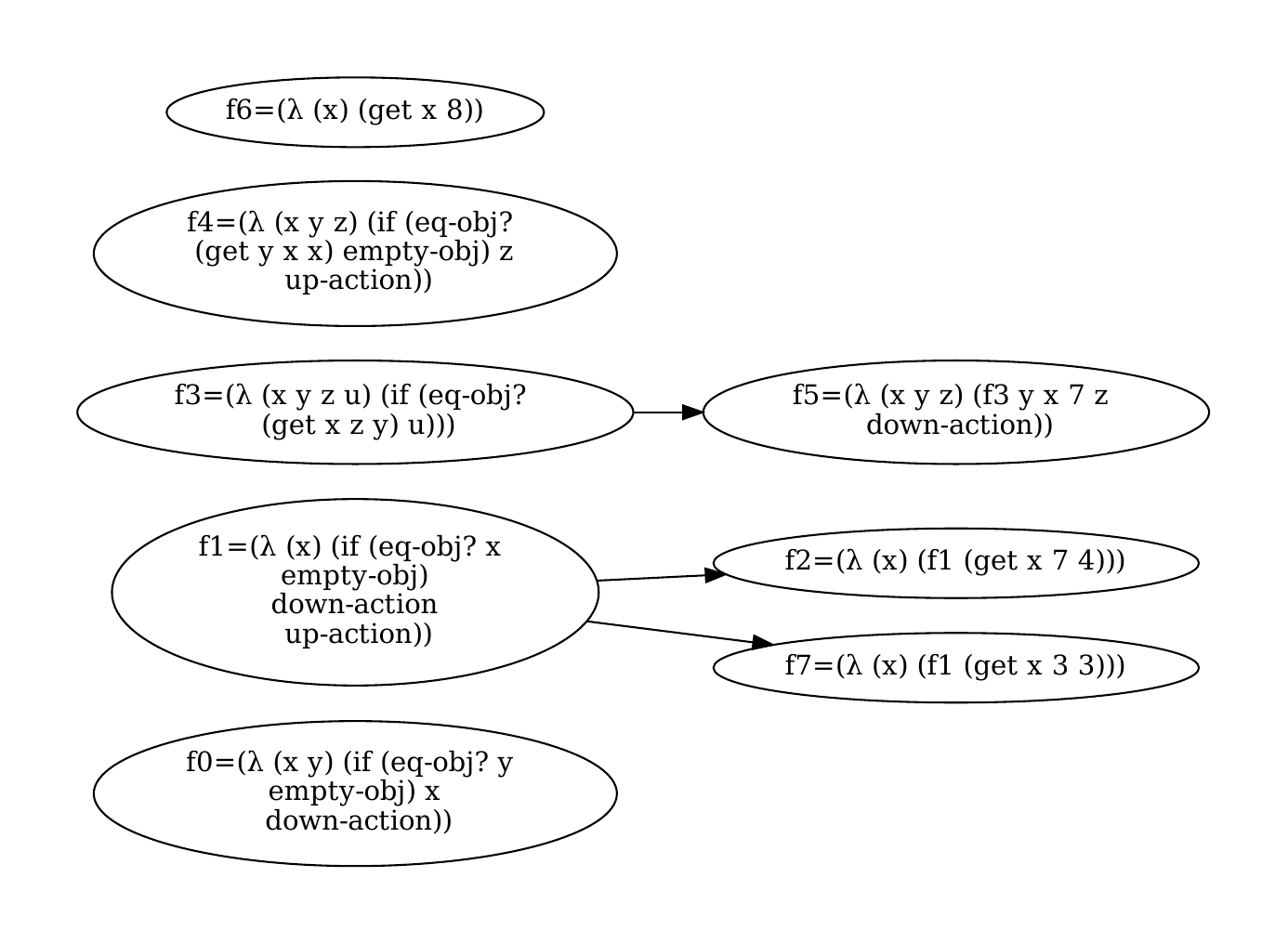}
  \caption{Asterix: The extracted functions from programs found by using LibT5.}
    \label{fig:t5-asterix-lib}
\end{figure}
\begin{figure}
  \centering
  \includegraphics[width=\linewidth]{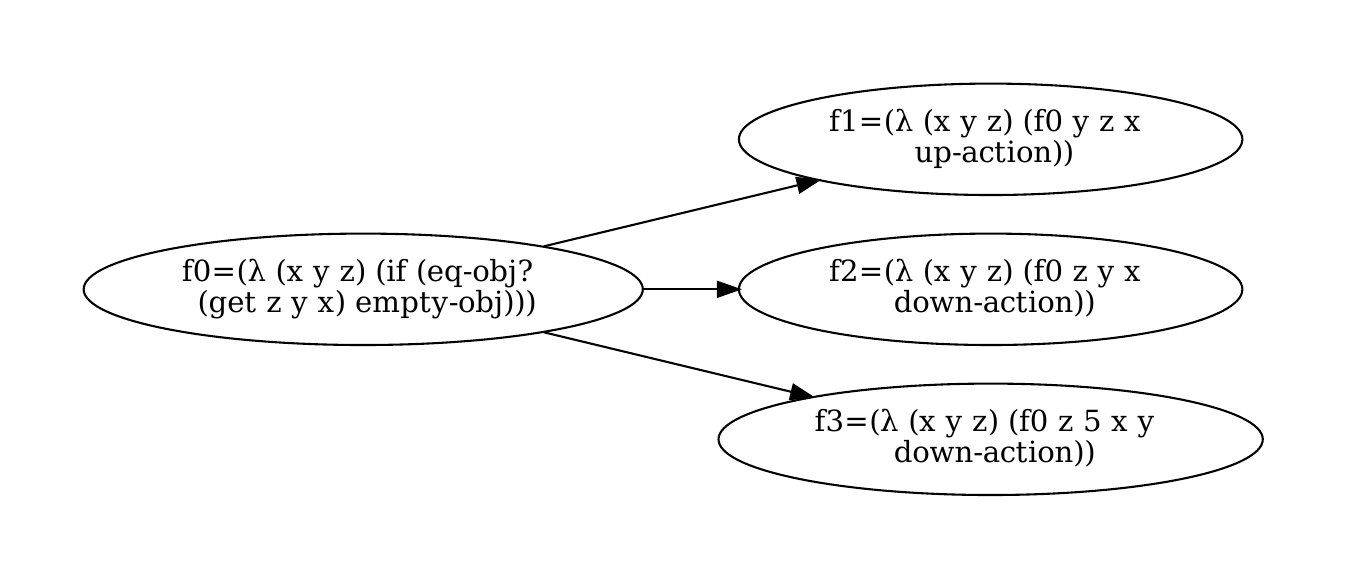}
  \caption{Asterix: The extracted functions from programs found by using DreamCoder.}
    \label{fig:ec-asterix-lib}
\end{figure}
\begin{figure}
  \centering
  \includegraphics[width=\linewidth]{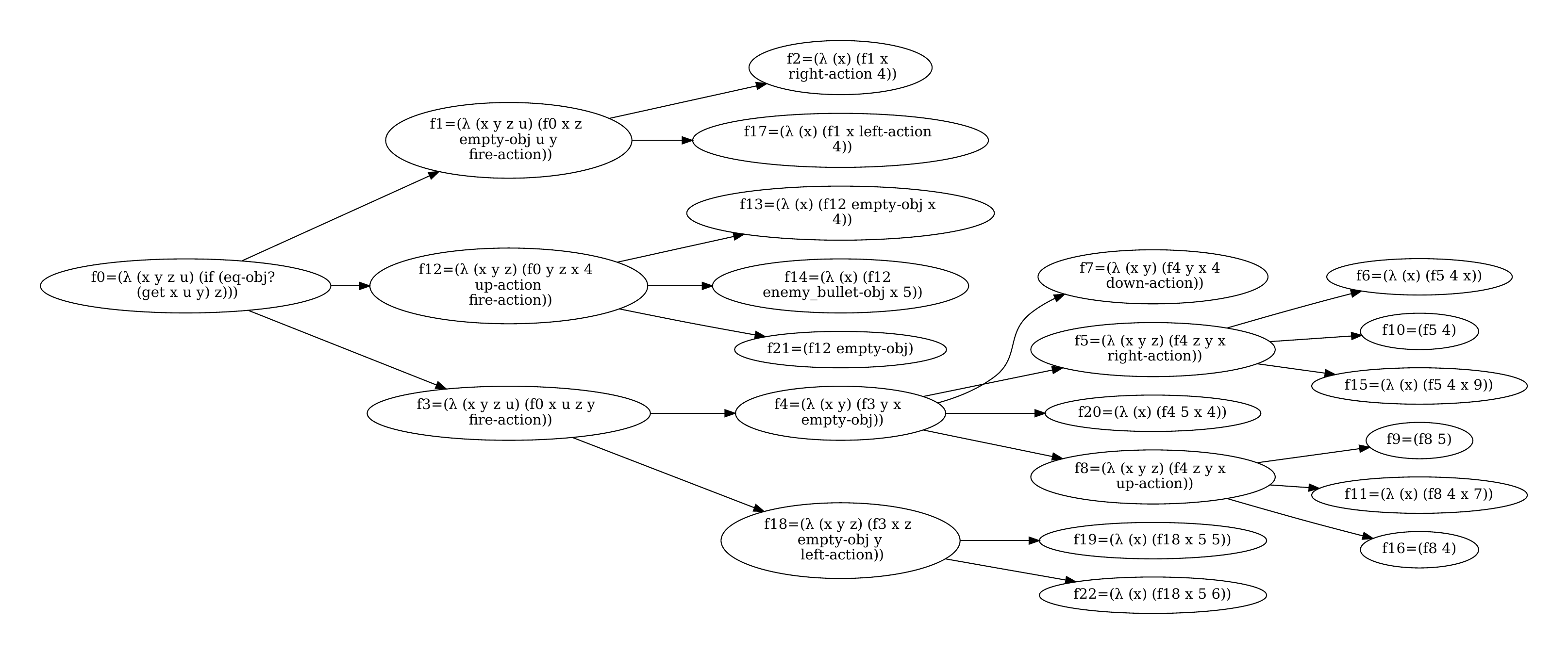}
  \caption{Space Invaders: The extracted functions from programs found by using LibT5 (zoom in for better visibility).}
    \label{fig:t5-space-invaders-lib}
\end{figure}
\begin{figure}
  \centering
  \includegraphics[width=\linewidth]{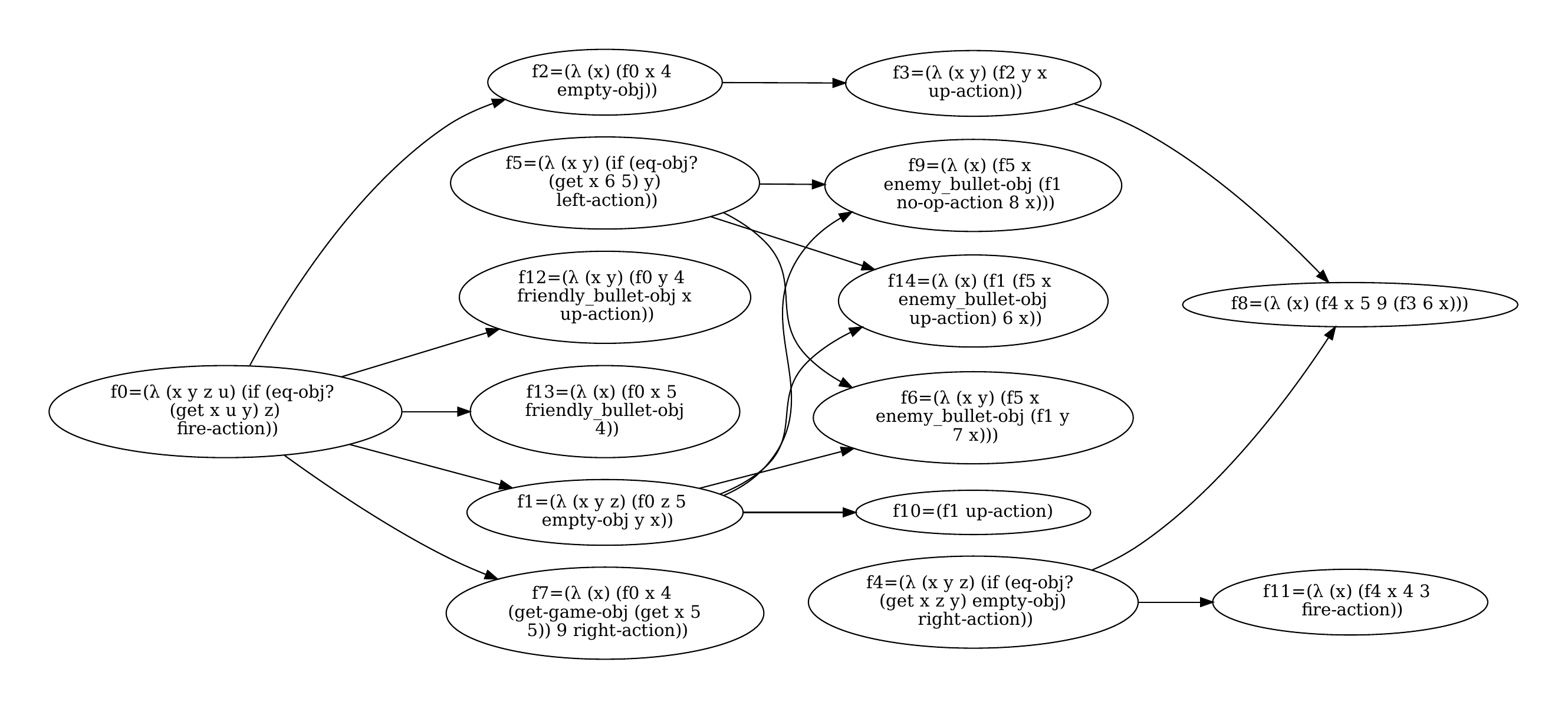}
  \caption{Space Invaders: The extracted functions from programs found by using DreamCoder (zoom in for better visibility).}
    \label{fig:ec-space-invaders-lib}
\end{figure}
\label{appendix:libs}
In this section, we present the full libraries created by DreamCoder and LibT5 for the three evaluated environments. Figure \ref{fig:t5-maze-lib} shows the library of LibT5 for the maze environment. For the Asterix environment both methods could not extract that many functions (see Figure \ref{fig:ec-asterix-lib} and \ref{fig:t5-asterix-lib}), we think that this shows the difficulty of the game compared to the maze environment. For Space Invaders, LibT5's library in Figure \ref{fig:t5-space-invaders-lib} shows a deeper hierarchical structure compared to DreamCoder's extracted functions in Figure \ref{fig:ec-space-invaders-lib}, but the evaluation has shown that a larger library is not always useful for the program synthesizer, especially for neural program synthesis.

\section*{Acknowledgements}
The work of S. M. was supported by the German Research Foundation under Grant MA 7111/7-1.

\bibliography{aaai23}
\end{document}